\documentclass[5p]{elsarticle}

\usepackage{amsmath}
\usepackage{amsfonts}
\usepackage{amssymb}
\usepackage{graphicx}%,epstopdf}
\usepackage{textcomp}
\usepackage{yhmath}
\usepackage{stmaryrd}
\usepackage{xspace}
\usepackage{mathtools}
\usepackage[usenames,dvipsnames]{color}
\DeclareMathOperator*{\argmax}{arg\,max}

\DeclarePairedDelimiter{\abs}{\vert}{\vert}
\usepackage[lined,boxed,commentsnumbered]{algorithm2e}
\usepackage{setspace}
\usepackage{hhline}
\usepackage{booktabs}
\usepackage{natbib}
\usepackage[colorlinks=true]{hyperref}

\journal{Engineering Applications of Artificial Intelligence}

\begin{document}
\emergencystretch 3em

\begin{frontmatter}

\title{
\begin{normalsize}
\begin{flushleft}
\copyright 2017. This manuscript version is made available under the CC-BY-NC-ND 4.0 license \url{http://creativecommons.org/licenses/by-nc-nd/4.0/}.\\
\medskip
This is a preprint version. The final version of this paper is available at \url{http://www.sciencedirect.com/science/article/pii/S0952197617301537} (DOI:10.1016/j.engappai.2017.07.005).\\
\end{flushleft}
\hrulefill
\color{white}
.\\
\end{normalsize}
\color{black}
\medskip
\medskip
\medskip
Particle Swarm Optimization for Generating Interpretable Fuzzy Reinforcement Learning Policies}
\author[tu,sie]{Daniel~Hein\corref{cor}}
\ead{daniel.hein@in.tum.de}
\author[az]{Alexander~Hentschel}
\author[sie]{Thomas~Runkler}
\author[sie]{Steffen~Udluft}

\address[tu]{Technische Universit\"at M\"uchen, Department of Informatics, Boltzmannstr. 3, 85748 Garching, Germany}
\address[sie]{Siemens AG, Corporate Technology, Otto-Hahn-Ring 6, 81739 Munich, Germany}
\address[az]{AxiomZen, 980 Howe St \#350, Vancouver, BC V6Z 1N9, Canada}

\cortext[cor]{Corresponding author}

\begin{abstract}
Fuzzy controllers are efficient and interpretable system controllers for continuous state and action spaces. 
To date, such controllers have been constructed manually or trained automatically either using expert-generated problem-specific cost functions or incorporating detailed knowledge about the optimal control strategy. 
Both requirements for automatic training processes are not found in most real-world reinforcement learning (RL) problems. 
In such applications, online learning is often prohibited for safety reasons because it requires exploration of the problem's dynamics during policy training.
We introduce a fuzzy particle swarm reinforcement learning (FPSRL) approach that can construct fuzzy RL policies solely by training parameters on world models that simulate real system dynamics. 
These world models are created by employing an autonomous machine learning technique that uses previously generated transition samples of a real system. 
To the best of our knowledge, this approach is the first to relate self-organizing fuzzy controllers to model-based batch RL. 
FPSRL is intended to solve problems in domains where online learning is prohibited, system dynamics are relatively easy to model from previously generated default policy transition samples, and it is expected that a relatively easily interpretable control policy exists.
The efficiency of the proposed approach with problems from such domains is demonstrated using three standard RL benchmarks, i.e., mountain car, cart-pole balancing, and cart-pole swing-up. 
Our experimental results demonstrate high-performing, interpretable fuzzy policies.
\end{abstract}


\end{frontmatter}

\section{Introduction}

This work is motivated by typical industrial application scenarios.
Complex industrial plants, like wind or gas turbines, have already been operated in the field for years.
For these plants, low-level control is realized by dedicated expert-designed controllers, which guarantee safety and stability.
Such low-level controllers are constructed with respect to the plant's subsystem dependencies which can be modeled by expert knowledge and complex mathematical abstractions, such as first principle models and finite element methods.
Examples for low-level controllers include self-organizing fuzzy controllers, which are considered to be efficient and interpretable~\citep{casillas:03} system controllers in control theory for decades~\citep{procyk:79, scharf:85, shao:88}.

However, we observed that high-level control is usually implemented by default control strategies, provided by best practice approaches or domain experts who are maintaining the system based on personal experience and knowledge about the system's dynamics.
One reason for the lack of autonomously generated real-world controllers is that modeling system dependencies for high-level control by dedicated mathematical representations is a complicated and often infeasible approach.
Further, modeling such representations by closed-form differentiable equations, as required by classical controller design, is even more complicated.
Since in many real-world applications such equations cannot be found, training high-level controllers has to be performed on reward samples from the plant. 
Reinforcement learning (RL)~\citep{sutton:98} is capable of yielding high-level controllers based solely on available system data.

Generally, RL is concerned with optimization of a policy for a system that can be modeled as a Markov decision process. 
This policy maps from system states to actions in the system. 
Repeatedly applying an RL policy generates a trajectory in the state-action space (Section~\ref{section:rl}).
Learning such RL controllers in a way that produces interpretable high-level controllers is the scope of this paper and the proposed approach.
Especially for real-world industry problems this is of high interest, since interpretable RL policies are expected to yield higher acceptance from domain experts than black-box solutions~\citep{maes:12}.

A fundamental difference between classical control theory and machine learning approaches, such as RL, lies in the way how these techniques address stability and reward function design.
In classical control theory, stability is the central property of a closed-loop controller. 
For example, Lyapunov stability theory analyzes the stability of a solution near a point of equilibrium. 
It is widely used to design controllers for nonlinear systems~\citep{lam:07}. 
Moreover, fault detection and robustness are of interest for fuzzy systems~\citep{yang:13,yang:14,yang:141}.
The problems addressed by classical fuzzy control theory, i.e., stability, fault detection, and robustness, make them well suited for serving as low-level system controllers.
For such controllers, reward functions specifically designed for the purpose of parameter training are essential.

In contrast, the second view on defining reward functions, which is typically applied in high-level system control, is to sample from a system's latent underlying reward dynamic and subsequently use this data to perform machine learning.
Herein, we apply this second view on defining reward functions, because RL is capable of utilizing sampled reward data for controller training. 
Note that the goal of RL is to find a policy that maximizes the trajectory's expected accumulated rewards, referred to as return value, without explicitly considering stability. 

Several approaches for autonomous training of fuzzy controllers have proven to produce remarkable results on a wide range of problems.
\citet{jang:93} introduced ANFIS, a fuzzy inference system implemented using an adaptive network framework. 
This approach has been frequently applied to develop fuzzy controllers. 
For example, ANFIS has been successfully applied to the cart-pole (CP) balancing problem~\citep{saifizul:06, hanafy:11, kharola:14}.
During the ANFIS training process, training data must represent the desired controller behavior, which makes this process a supervised machine learning approach. 
However, the optimal controller trajectories are unknown in many industry applications.

\citet{feng:051,feng:05} applied particle swarm optimization (PSO) to generate fuzzy systems to balance the CP system and approximate a nonlinear function 
Debnath et al.\ optimized Gaussian membership function parameters for nonlinear problems and showed that parameter tuning is much easier with PSO than with conventional methods because knowledge about the derivative and complex mathematical equations are not required~\citep{debnath:13}. 
\citet{kothandaraman:12} applied PSO to tune adaptive neuro fuzzy controllers for a vehicle suspension system. 
However, similar to ANFIS, the PSO fitness functions in all these contributions have been dedicated expert formulas or mean-square error functions that depend on correctly classified samples.

To the best of our knowledge, self-organizing fuzzy rules have never been combined with a model-based batch RL approach. 
In the proposed fuzzy particle swarm reinforcement learning (FPSRL) approach, different fuzzy policy parameterizations are evaluated by testing the resulting policy on a world model using a Monte Carlo method~\citep{sutton:98}. 
The combined return value of a number of action sequences is the fitness value that is maximized iteratively by the optimizer.

In batch RL, we consider applications where online learning approaches, such as classical temporal-difference learning~\citep{sutton:88}, are prohibited for safety reasons, since these approaches require exploration of system dynamics. 
In contrast, batch RL algorithms generate a policy based on existing data and deploy this policy to the system after training. 
In this setting, either the value function or the system dynamics is trained using historic operational data comprising a set of four-tuples of the form (\textit{observation}, \textit{action}, \textit{reward}, \textit{next observation}), which is referred to as a data batch.
Research from the past two decades~\citep{gordon:95,ormoneit:02,lagoudakis:03,ernst:05} suggests that such batch RL algorithms satisfy real-world system requirements, particularly when involving neural networks (NNs) modeling either the state-action value function~\citep{riedmiller:051,riedmiller:05,nrr:07,schneegass:07,riedmiller:09} or system dynamics~\citep{bakker:04,schafer:08,depeweg:16}. 
Moreover, batch RL algorithms are data-efficient~\citep{riedmiller:051,schaefer:07} because batch data is utilized repeatedly during the training phase.

FPSRL is a model-based approach, i.e., training is conducted on an environment approximation referred to as world model.
Generating a world model from real system data in advance and training a fuzzy policy offline using this model has several advantages. 
(1) In many real-world scenarios, data describing system dynamics is available in advance or is easily collected. 
(2) Policies are not evaluated on the real system, thereby avoiding the detrimental effects of executing a bad policy. 
(3) Expert-driven reward function engineering yielding a closed-form differentiable equation utilized during policy training is not required, i.e., it is sufficient to sample from the system's reward function and model the underlying dependencies using supervised machine learning.

The remainder of this paper is organized as follows.
The methods employed in our framework are reviewed in Sections \ref{section:rl}--\ref{section:pso}.
Specifically, the problem of finding policies via RL is formalized as an optimization task. 
In addition, we review Gaussian-shaped membership functions and describe the proposed parameterization approach. 
Finally, PSO, an optimization heuristic we use for searching for optimal policy parameters, and its different extensions are presented. 
An overview of how the proposed FPSRL approach is derived from different methods is given in Section~\ref{section:psrl}.

Experiments using three benchmark problems, i.e., the mountain car (MC) problem, the CP balancing (CPB) task, and the more complex CP swing-up (CPSU) challenge, are described in Section~\ref{section:experiments}.
In this section, we also explain the setup process of the world models and introduce the applied fuzzy policies.

Experimental results are discussed in Section~\ref{section:results}. 
The results demonstrate that the proposed FPSRL approach can solve the benchmark problems and is human-readable and understandable. 
To benchmark FPSRL, we compare the obtained results to those of neural fitted Q iteration (NFQ)~\citep{riedmiller:051,riedmiller:05}, an  established RL technique. 
Note that this technique was chosen to describe the advantages and limitations of the proposed method compared to a well-known, widely available standard algorithm.
\section{Model-based Reinforcement Learning}
\label{section:rl}

In biological learning, an animal interacts with its environment and attempts to find action strategies to maximize its perceived accumulated reward. 
This notion is formalized in RL, an area of machine learning where the acting agent is not explicitly told which actions to implement.
Instead, the agent must learn the best action strategy from the observed environment's responses to the agent's actions. 
For the most common (and most challenging) RL problems, an action affects both the next reward and subsequent rewards~\citep{sutton:98}. 
Examples for such delayed effects are nonlinear change in position when a force is applied to a body with mass or delayed heating in a combustion engine.

In the RL formalism, the agent interacts with the target system in discrete time steps, $t=0,1,2,\ldots$. 
At each time step, the agent observes the system's state $\mathbf{s}_t \in \mathcal S$ and applies an action $\mathbf{a}_t \in \mathcal A$, where $\mathcal S$ is the state space and $\mathcal A$ is the action space. 
Depending on $\mathbf{s}_t$ and $\mathbf{a}_t$, the system transitions to a new state and the agent receives a real-value reward $r_{t+1} \in \mathbb{R}$. 
Herein, we focus on deterministic systems where state transition $g$ and reward $r$ can be expressed as functions $g:\mathcal S \times \mathcal A \rightarrow \mathcal S$ with $g(\mathbf{s}_t,\mathbf{a}_t)=\mathbf{s}_{t+1}$ and $r:\mathcal S \times \mathcal A \times \mathcal S \rightarrow \mathbb{R}$ with $r(\mathbf{s}_t,\mathbf{a}_t,\mathbf{s}_{t+1})=r_{t+1}$, respectively. 
The desired solution to an RL problem is an action strategy, i.e., a policy, that maximizes the expected cumulative reward, i.e., return $\mathcal{R}$.

In our proposed setup, the goal is to find the best policy among a set of policies that is spanned by a parameter vector $\mathbf{x\in \mathcal X}$. 
Herein, a policy corresponding to a particular parameter value $\mathbf{x}$ is denoted by $\pi[\mathbf{x}]$. 
For state $\mathbf{s}_t$, the policy outputs action $\pi[\mathbf{x}](\mathbf{s}_t)=\mathbf{a}_t$. 
The policy's performance when starting from $\mathbf{s}_t$ is measured by the return $\mathcal{R}$, i.e., the accumulated future rewards obtained by executing the policy. 
To account for increasing uncertainties when accumulating future rewards, the reward $r_{t+k}$ for $k$ future time steps is weighted by $\gamma^k$, where $\gamma\in[0,1]$.
Furthermore, adopting a common approach, we include only a finite number of $T>1$ future rewards in the return~\citep{sutton:98}, which is expressed as follows:
\begin{equation}\label{eq:return}
  \begin{aligned}
    \mathcal R (\mathbf{s}_t,\pi[\mathbf{x}]) & = \sum_{k=0}^{T-1}\gamma^kr(\mathbf{s}_{t+k},\pi[\mathbf{x}](\mathbf{s}_{t+k}),\mathbf{s}_{t+k+1}), \\
    \textnormal{with}\quad \mathbf{s}_{t+k+1} & = g(\mathbf{s}_{t+k},\mathbf{a}_{t+k}).
  \end{aligned}
\end{equation}
The discount factor $\gamma$ is selected such that, at the end of the time horizon $T$, the last reward accounted for is weighted by $q\in[0,1]$, yielding $\gamma=q^{1/(T-1)}$.
The overall state-independent policy performance $\mathcal{F}(\mathbf{x})$ is obtained by averaging over all starting states $s_t \in \mathcal S$ using their respective probabilities $w_{\mathbf{s}_t}$ as weight factors. 
Thus, optimal solutions to the RL problem are $\pi[\mathbf{x}]$ with
\begin{equation}\label{eq:fitness_function}
    \mathbf{\hat{x}} \in \argmax_{\mathbf{x\in \mathcal X}}\mathcal{F}(\mathbf{x}), \quad\text{with}\quad \mathcal{F}(\mathbf{x})=\sum_{\mathbf{s}_t\in \mathcal S}w_{\mathbf{s}_t}\mathcal R (\mathbf{s}_t,\pi[\mathbf{x}]).
\end{equation}
In optimization terminology, the policy performance function $\mathcal{F}(\mathbf{x})$ is referred to as a fitness function.

For many real-world problems, the cost of executing a potentially bad policy is prohibitive. 
This is why, e.g., pilots learn flying using a flight simulator instead of real aircraft. 
Similarly, in model-based RL~\citep{busoniu:10}, the real-world state transition function $g$ is approximated using a model $\tilde g$, which can be a first principle model or created from previously gathered data. 
By substituting $\tilde{g}$ in place of the real-world state transition function $g$ in Eq.~\eqref{eq:return}, we obtain a model-based approximation $\mathcal{\tilde F}(\mathbf{x})$ of the true fitness function Eq.~\eqref{eq:fitness_function}. 
In this study, we employ models based on NNs. 
However, the proposed method can be extended to other models, such as Bayesian NNs~\citep{depeweg:16} and Gaussian process models~\citep{rasmussen:06}. 

\section{Fuzzy Rules}
\label{section:fr}

Fuzzy set theory was introduced by \citet{zadeh:65}. 
Based on this theory, \citet{mamdani:75} introduced a so-called fuzzy controller specified by a set of linguistic if-then rules whose membership functions can be activated independently and produce a combined output computed by a suitable defuzzification function.

In a $D$-inputs-single-output system with $C$ rules, a fuzzy rule $R^{(i)}$ can be expressed as follows:
\begin{equation}
    R^{(i)}: \text{ IF }\mathbf{s}\text{ is } m^{(i)} \text{ THEN }o^{(i)}, \quad \text{with }i\in\{1,\dotsc,C\}, 
\end{equation}
where $\mathbf{s}\in \mathbb{R}^{D}$ denotes the input vector (the environment state in our setting), $m^{(i)}$ is the membership of a fuzzy set of the input vector in the premise part, and $o^{(i)}$ is a real number in the consequent part.

In this paper, we apply Gaussian membership functions~\citep{wang:92}.
This very popular type of membership function yields smooth outputs, is local but never produces zero activation, and  forms a multivariate Gaussian function by applying the product over all membership dimensions.
We define the membership function of each rule as follows:
\begin{equation}
    m^{(i)}(\mathbf{s})=\text{m}[\mathbf{c}^{(i)},\mathbf{\sigma}^{(i)}](\mathbf{s})=\prod^{D}_{j=1}\exp\left\{-\frac{(c_{j}^{(i)}-s_j)^2}{2{\sigma_{j}^{(i)}}^2}\right\},
    \label{gaussian_membership}
\end{equation}
where $m^{(i)}$ is the i-th parameterized Gaussian $\text{m}(\mathbf{c},\mathbf{\sigma})$ with its center at $\mathbf{c}^{(i)}$ and width $\mathbf{\sigma}^{(i)}$.

The parameter vector $\mathbf{x}\in\mathcal X$, where $\mathcal X$ is the set of valid Gaussian fuzzy parameterizations, is of size $d=(2D+1)\cdot C + 1$ and contains
\begin{align}
\label{eq:x_vector}
    \begin{split}
    \mathbf{x}=( & c_{1}^{(1)},c_{2}^{(1)},\dotsc,c_{D}^{(1)},\sigma_{1}^{(1)},\sigma_{2}^{(1)},\dotsc,\sigma_{D}^{(1)},o^{(1)},\\
    & c_{1}^{(2)},c_{2}^{(2)},\dotsc,c_{D}^{(2)},\sigma_{1}^{(2)},\sigma_{2}^{(2)},\dotsc,\sigma_{D}^{(2)},o^{(2)},\dotsc,\\
    & c_{1}^{(C)},c_{2}^{(C)},\dotsc,c_{D}^{(C)},\sigma_{1}^{(C)},\sigma_{2}^{(C)},\dotsc,\sigma_{D}^{(C)},o^{(C)},\alpha).
    \end{split}
\end{align}

The output is determined using the following formula: 
\begin{equation}
    \pi[\mathbf{x}](\mathbf{s})=\tanh\left(\alpha\cdot\frac{\sum_{i=1}^{C}m^{(i)}(\mathbf{s})\cdot o^{(i)}}{\sum_{i=1}^{C}m^{(i)}(\mathbf{s})}\right),
    \label{eq:defuzzifier}
\end{equation}
where the hyperbolic tangent limits the output to between -1 and 1, and parameter $\alpha$ can be used to change the slope of the function.
\section{Particle Swarm Optimization}
\label{section:pso}

The PSO algorithm is a population-based, non-convex, stochastic optimization heuristic. Generally, PSO can operate on any search space that is a bounded sub-space of a finite-dimensional vector space~\citep{kennedy:95}.

The position of each particle in the swarm represents a potential solution of the given problem. 
The particles fly iteratively through the multidimensional search space, which is referred to as the fitness landscape. 
After each movement, each particle receives a fitness value for its new position.
This fitness value is used to update a particle's own velocity vector and the velocity vectors of all particles in a certain neighborhood.

At each iteration $p$, particle $i$ remembers the best local position $\mathbf{y}_i(p)$ it has visited so far (including its current position). 
Furthermore, particle $i$ knows the neighborhood's best position
\begin{equation}
\label{neighborhood_best}
	\hat{\mathbf{y}}_i(p)\in\argmax_{\mathbf{z}\in \{\mathbf{y}_j(p)\mid j\in\mathcal{N}_i\}}\mathcal{F}(\mathbf{z}),
\end{equation}
found so far by any one particle in its neighborhood $\mathcal{N}_i$ (including itself). 
The neighborhood relations between particles are determined by the swarm's population topology and are generally fixed, irrespective of the particles' positions. 
Note that a ring topology~\citep{eberhart:96} is used in the experiments described in Section~\ref{section:experiments}.

Let $\mathbf{x}_i(p)$ denote the position of particle $i$ at iteration $p$. 
Changing the position of a particle in each iteration is achieved by adding the velocity vector $\mathbf{v}_i(p)$ to the particles position vector
\begin{equation}
\label{position_update}
\mathbf{x}_i(p+1)=\mathbf{x}_i(p)+\mathbf{v}_i(p+1),
\end{equation}
where $\mathbf{x}_i(0)\sim U(\mathbf{x}_{\text{min}},\mathbf{x}_{\text{max}})$ is distributed uniformly.

The velocity vector contains both a cognitive component and a social component that represent the attraction to the given particle's best position and the neighborhood's best position, respectively. 
The velocity vector is calculated as follows:
\begin{align}
\begin{split}
    \label{basic_pso}
    v_{ij}(p+1)=&wv_{ij}(p)+\underbrace{c_1r_{1j}(p)[y_{ij}(p)-x_{ij}(p)]}_{\text{cognitive component}}\\
    &+\underbrace{c_2r_{2j}(p)[\hat{y}_{ij}(p)-x_{ij}(p)]}_{\text{social component}},
\end{split}
\end{align}
where $w$ is the inertia weight factor, $v_{ij}(p)$ and $x_{ij}(p)$ are the velocity and position of particle $i$ in dimension $j$, and $c_1$ and $c_2$ are positive acceleration constants used to scale the contribution of the cognitive and social components $y_{ij}(p)$ and $\hat{y}_{ij}(p)$, respectively. 
The factors $r_{1j}(p)$ and $r_{2j}(p)\sim U(0,1)$ are random values sampled from a uniform distribution to introduce a stochastic element to the algorithm.

The best position of a particle for a maximization problem at iteration $p$ is calculated as follows:
\begin{equation}
\label{personal_best}
\mathbf{y}_i(p)=\begin{cases}
\mathbf{x}_i(p), & \text{if }\mathcal{F}(\mathbf{x}_i (p))>\mathcal{F}(\mathbf{y}_i (p-1))\\
\mathbf{y}_i(p-1), & \text{else},
\end{cases}
\end{equation}
where in our framework $\mathcal{F}$ is the fitness function given in Eq.~\eqref{eq:fitness_function} and the particle positions represent the policy's parameters $\mathbf{x}$ from Eq.~\eqref{eq:x_vector}.

Pseudocode for the PSO algorithm applied in our experiments (Section~\ref{section:experiments}) is provided in \ref{appendix:algorithm}.
\section{Fuzzy Particle Swarm Reinforcement Learning}
\label{section:psrl}

The basis for the proposed FPSRL approach is a data set $\mathcal{D}$ that contains state transition samples gathered from a real system. 
These samples are represented by tuples $(\mathbf{s},\mathbf{a},\mathbf{s}',r)$, 
where, in state $\mathbf{s}$, action $\mathbf{a}$ was applied and resulted in state transition to $\mathbf{s}'$ and yielded reward $r$.
$\mathcal{D}$ can be generated using any (even a random) policy prior to policy training.

Then, we generate world models $\tilde g$ with inputs $(\mathbf{s},\mathbf{a})$ to predict $\mathbf{s}'$, using data set $\mathcal{D}$. 
It is advantageous to learn the differences of the state variables and train a single model per state variable separately to yield better approximative quality:
\begin{align*}
    \Delta s'_1 & = \tilde g_{s_1}(s_1,s_2,\dots,s_m,\mathbf{a})\\
    \Delta s'_2 & = \tilde g_{s_2}(s_1,s_2,\dots,s_m,\mathbf{a})\\
    \dots\\
    \Delta s'_m & = \tilde g_{s_m}(s_1,s_2,\dots,s_m,\mathbf{a}).\\
\end{align*}
Then, the resulting state is calculated according to $\mathbf{s}'=(s_1+\Delta s'_1,s_2+\Delta s'_2,\dots,s_m+\Delta s'_m)$.
The reward is also given in $\mathcal{D}$; thus, the reward function can be approximated using $r=\tilde r(\mathbf{s},\mathbf{a},\mathbf{s}')$.

For the next FPSRL step, an assumption about the number of rules per policy is required. 
In our experiments, we started with a minimal rule set for each benchmark and calculated the respective performances. 
Then, we increased the number of rules and compared the performance with those of the policies with fewer rules. 
This process was repeated until performance with respect to the dynamic models was satisfactory.
An intuitive representation of the maximal achievable policy performance given a certain discount factor with respect to a particular model can be computed by adopting a trajectory optimization technique, such as PSO-P~\citep{hein:16}, prior to FPSRL training.

\begin{figure*}[tp]
	\centering
	\includegraphics[width=5.0in]{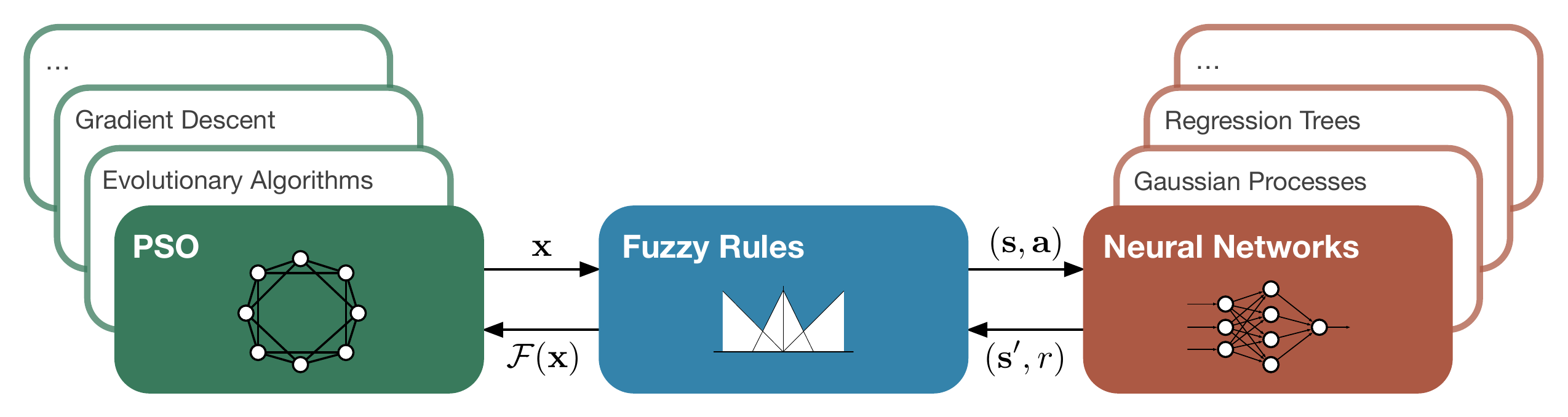}
	\caption{
		Schematic visualization of the proposed FPSRL approach. 
		From left to right: 
		PSO evaluates parameter vectors $\mathbf{x}$ of a predefined fuzzy rule representation $\pi[\mathbf{x}]$. 
		For each given set of parameters, a model-based RL evaluation is performed by first computing an action vector $\mathbf{a}=\pi[\mathbf{x}](\mathbf{s})$ for state $\mathbf{s}$ (Eq.~\eqref{eq:defuzzifier}). 
		Then, the approximative performance of this tuple is computed by predicting both the resulting state $\mathbf{s}'$ and the transition's reward $r$ using NNs.
		Repeating this procedure for state $\mathbf{s}'$ and its successor states generates an approximative  trajectory through the state space.
		Accumulating the rewards using Eq.~\eqref{eq:return}, the return $\mathcal{R}$ is computed for each state, which is eventually used to compute the fitness value $\tilde{\mathcal{F}}(\mathbf{x})$, which drives the swarm to high performance policy parameterizations (Eq.~\eqref{eq:fitness_function}).
		Alternative techniques that could replace PSO and NNs are presented in the background.
	}
	\label{psrl}
\end{figure*}

During optimization, each particle's position $\mathbf{x}$ in the PSO represents a parameterization of the fuzzy policy $\pi[\mathbf{x}]$. 
The fitness $\mathcal{\tilde F}$ of a particle is calculated by generating trajectories using the world model $\tilde g$ starting from a fixed set of initial benchmark states (Section~\ref{section:rl}).
A schematic representation of the proposed FPSRL framework is given in Fig.~\ref{psrl}.

Note that we present the results of FPSRL using NNs as world models and PSO as the optimization technique.
In the considered problem domain, i.e., continuous, smooth, and deterministic system dynamics, NNs are known to serve as adequate world models. 
Given a batch of previously generated transition samples, the NN training process is data-efficient and  training errors are excellent indicators of how well the model will perform in model-based RL training.
Nevertheless, for different problem domains, alternative types of world models might be preferable. 
For example, Gaussian processes~\citep{rasmussen:06} provide a good approximation of the mean of the target value, and  this technique indicates the level of confidence about this prediction.
This feature may be of value for stochastic system dynamics.
A second alternative modeling technique is the use of regression trees~\citep{breiman:84}.
While typically lacking data efficiency, regression tree predictions are less affected by nonlinearities perceived by  system dynamics because they do not rely on a closed-form functional approximation.

We employed PSO in our study because the population-based optimizer does not require any gradient information about its fitness landscape.
PSO utilizes neighborhood information to systematically search for valuable regions in the search space. 
Note that gradient-descent based methods or evolutionary algorithms are alternative techniques.
\section{Experiments}
\label{section:experiments}

\subsection{Mountain Car}

\begin{figure}[!t]
    \centering
    \includegraphics[width=3.5in]{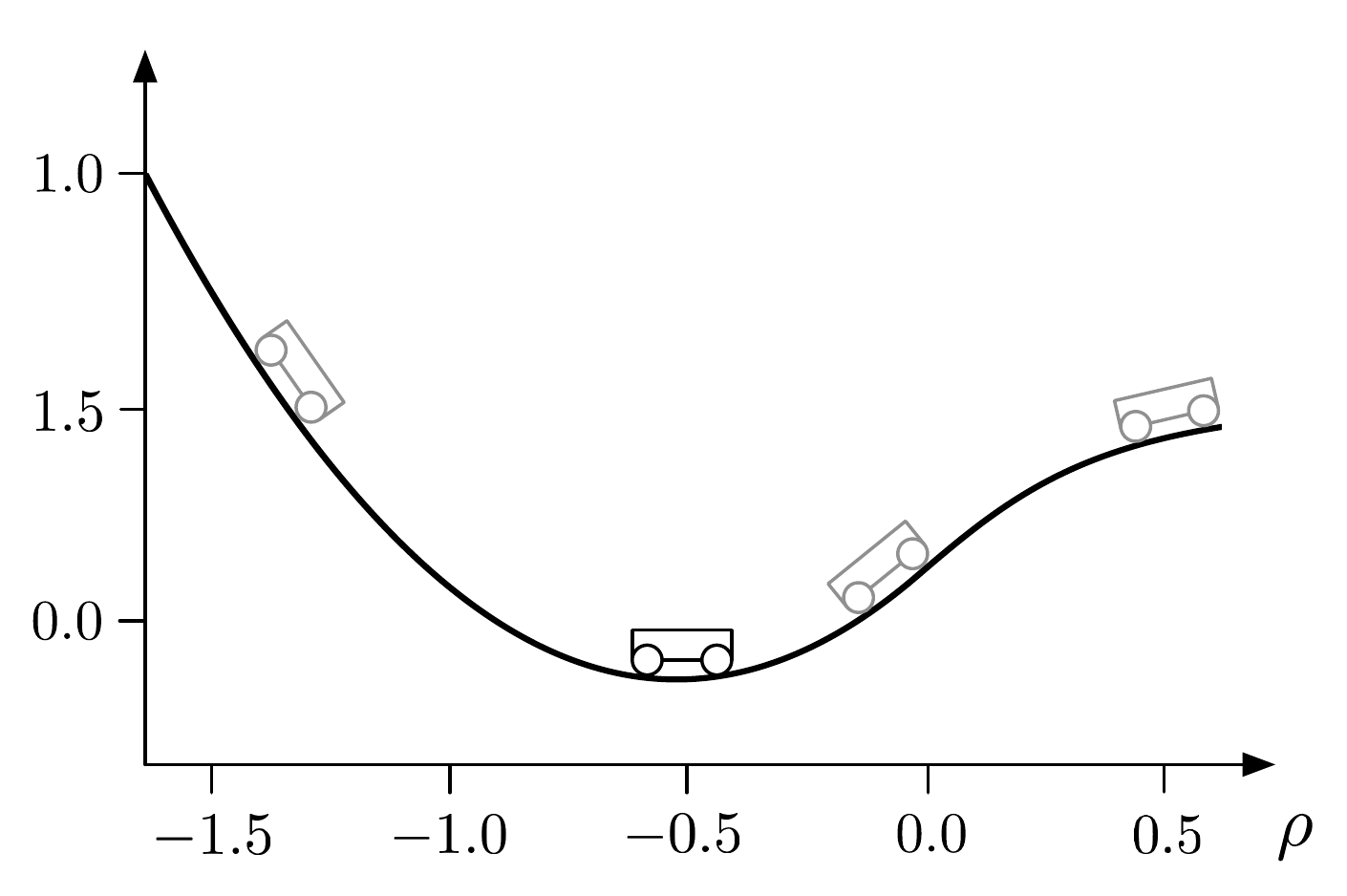}
    \caption{
    	Mountain car benchmark.
    	The task is to first build up momentum by driving to the left in order to subsequently reach the top of the hill on the right at $\rho=0.6$.
    }
    \label{mountain_car}
\end{figure}

In the MC benchmark, an underpowered car must be driven to the top of a hill (Fig.~\ref{mountain_car}). 
This is achieved by building sufficient potential energy by first driving in the direction opposite to the final direction. 
The system is fully described by the two-dimensional state space $\mathbf{s}=(\rho,\dot{\rho})$ representing the cars position $\rho$ and velocity $\dot{\rho}$.

We conducted MC experiments using the freely available $CLS^2$ software ('clsquare')\footnote{http://ml.informatik.uni-freiburg.de/research/clsquare.}, which is an RL benchmark system that applies the Runge-Kutta fourth-order method to approximate closed loop dynamics.
The task for the RL agent is to find a sequence of force actions $a_t,a_{t+1},a_{t+2},\ldots\in[-1,1]$ that drive the car up the hill, which is achieved when reaching position $\rho\geq0.6$.

At the start of each episode, the car's position is initialized in the interval $[-1.2,0.6]$. 
The agent receives a reward of 
\begin{equation}
	r(\mathbf{s}')=
	\begin{cases}
		0, & \text{if }\rho'\geq0.6,\\
		-1, & \text{otherwise},
	\end{cases}
\end{equation}
subsequent to each state transition $\mathbf{s}'=g(\mathbf{s},a)$. 
When the car reaches the goal position, i.e., $\rho\geq0.6$, its position becomes fixed, the velocity becomes zero, and the agent perceives the maximum reward in each following time step regardless of the applied actions.

\subsection{Cart-pole Balancing}

\begin{figure}[!t]
    \centering
    \includegraphics[width=3.5in]{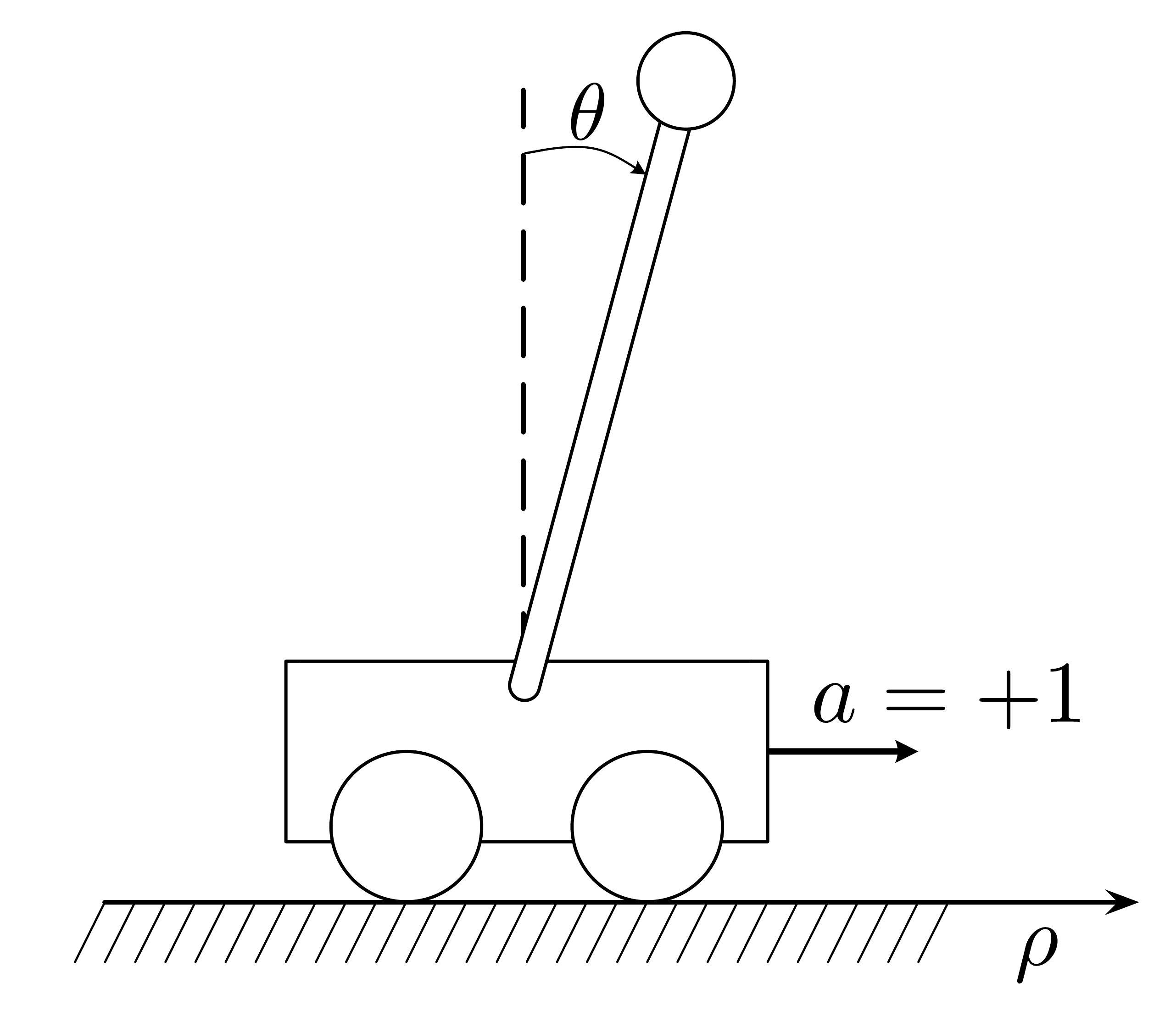}
    \caption{
    	Cart-pole benchmark.
    	The task is to balance the pole around $\theta=0$ while moving the cart to position $\rho=0$ by applying positive or negative force to the cart.
    }
    \label{cart_pole}
\end{figure}

The CP experiments described in the following two sections were also conducted using the $CLS^2$ software.
The objective of the CPB benchmark is to apply forces to a cart moving on a one-dimensional track to keep a pole hinged to the cart in an upright position (Fig.\ \ref{cart_pole}). 
Here, the four Markov state variables are the pole angle $\theta$, the pole angular velocity $\dot\theta$, the cart position $\rho$, and the cart velocity $\dot\rho$. 
These variables describe the Markov state completely, i.e., no additional information about the system's past behavior is required. 
The task for the RL agent is to find a sequence of force actions $a_t,a_{t+1},a_{t+2},\ldots$  that prevent the pole from falling over~\citep{fantoni:02}.

In the CPB task, the angle of the pole and the cart's position are restricted to intervals of $[-0.7,0.7]$ and $[-2.4,2.4]$ respectively. 
Once the cart has left the restricted area, the episode is considered a failure, i.e., velocities become zero, the cart's position and pole's angle become fixed, and the system remains in the failure state for the rest of the episode. 
The RL policy can apply force actions on the cart from $-10$~N to $+10$~N in time intervals of $0.025$~s.

The reward function for the balancing problem is given as follows:
\begin{equation}
	r(\mathbf{s}')=
	\begin{cases}
		0.0, & \text{if }\abs{\theta'}<0.25\\& \text{and } \abs{\rho'}<0.5,\\
		-1.0, & \text{if }\abs{\theta'}>0.7\\& \text{or } \abs{\rho'}>2.4,\\
		-0.1, & \text{otherwise}.
	\end{cases}
\end{equation}

Based on this reward function, the primary goal of the policy is to avoid reaching the failure state. 
The secondary goal is to drive the system to the goal state region where $r=0$ and keep it there for the rest of the episode.

Since the CP problem is symmetric around $\mathbf{s}=(\theta,\dot\theta,\rho,\dot\rho)=(0,0,0,0)$, an optimal action $a_t$ for state $(\theta,\dot\theta,\rho,\dot\rho)$ corresponds to an optimal action $-a_t$ for state $(-\theta,-\dot\theta,-\rho,-\dot\rho)$. 
Thus, the parameter search process can be simplified. 
It is only necessary to search for optimal parameters for one half of the fuzzy policy rules. 
The other half of the parameter sets can be constructed by negating the membership functions' mean parameters $c^{(i)}_j$ and the respective output values $o^{(i)}$ of the policy's components. 
Note that the membership function span width of the fuzzy rules (parameter $\sigma^{(i)}_j$ in Eq.~\eqref{gaussian_membership}) is not negated because the membership functions must preserve their shapes.

\subsection{Cart-pole Swing-up}

The CPSU benchmark is based on the same system dynamics as the CPB benchmark. 
In contrast to the CPB benchmark, the position of the cart and the angle of the pole are not restricted. 
Consequently, the pole can swing through, which is an important property of CPSU. 
Since the pole's angle is initialized in the full interval of $[-\pi,\pi]$, it is often necessary for the policy to swing the pole several times from side to side to gain sufficient energy to erect the pole and receive the highest reward.

In the CPSU setting, the policy can apply actions from $-30$~N to $+30$~N on the cart. 
The reward function for the problem is given as follows: 
\begin{equation}
	r(\mathbf{s}')=
	\begin{cases}
		0, & \text{if }\abs{\theta'}<0.5 \\& \text{and } \abs{\rho'}<0.5,\\
		-1, & \text{otherwise}.
	\end{cases}
\end{equation}
This is similar to the reward function for the CPB benchmark but does not contain any penalty for failure states, which terminate the episode when reached.

\subsection{Neural Network World Models}

We conducted policy training on NN world models, which yielded approximative fitness functions $\tilde {\mathcal{F}}(\mathbf{x})$ (Section~\ref{section:rl}).
For these experiments, we created one NN for each state variable. 
Prior to training, the respective data sets were split into blocks of 80\%, 10\%, and 10\% (training, validation and generalization sets, respectively).
While the weight updates during training were computed by utilizing the training sets, the weights that performed best given the validation sets were used as training results.
Finally, those weights were evaluated using the generalization sets to rate the overall approximation quality on unseen data.

The MC NNs were trained with data set $\mathcal{D}_{\text{MC}}$ containing tuples $(\mathbf{s},a,g(\mathbf{s},a),r)$ from trajectories generated by applying random actions on the benchmark dynamics. 
The start states for these trajectories were uniformly sampled as $\mathbf{s}=(\rho,\dot{\rho})\in[-1.2,0.6]\times\{0\}$, i.e., at a random position on the track with zero velocity. 
The following three NNs were trained to approximate the MC task:
\begin{align*}
    \Delta \rho_{t+1} & = \tilde g_{\rho}(\rho_t,{\dot\rho}_t,a_t),\\
    \Delta \dot\rho_{t+1} & = \tilde g_{\dot\rho}(\rho_t,{\dot\rho}_t,a_t),\\
    r_{t+1} & = \tilde r(\mathbf{s}_{t},a_t,\mathbf{s}_{t+1}),\\
    \text{with }\mathbf{s}_{t+1} &= (\rho_t+\Delta \rho_{t+1},\dot{\rho}_t+\Delta \dot\rho_{t+1}).
\end{align*}

Similarly, for the CP dynamic model state $\mathbf{s}_{t}=(\theta_t,{\dot\theta}_t,\rho_t,{\dot\rho}_t)$ we created the following four networks:
\begin{align*}
    \Delta \theta_{t+1} & = \tilde g_{\theta}(\theta_t,{\dot\theta}_t,\rho_t,{\dot\rho}_t,a_t)\\
    \Delta \dot\theta_{t+1} & = \tilde g_{\dot\theta}(\theta_t,{\dot\theta}_t,\rho_t,{\dot\rho}_t,a_t)\\
    \Delta \rho_{t+1} & = \tilde g_{\rho}(\theta_t,{\dot\theta}_t,\rho_t,{\dot\rho}_t,a_t)\\
    \Delta \dot\rho_{t+1} & = \tilde g_{\dot\rho}(\theta_t,{\dot\theta}_t,\rho_t,{\dot\rho}_t,a_t).
\end{align*}
An approximation of the next state is given by the following formula:
\begin{equation}
	\mathbf{s}_{t+1}=(\theta_t+\Delta \theta_{t+1},{\dot\theta}_t+\Delta \dot\theta_{t+1},\rho_t+\Delta \rho_{t+1},{\dot\rho}_t+\Delta \dot\rho_{t+1}).
\end{equation}
The result of this formula can subsequently be used to approximate the state transition's reward by
\begin{equation}
 	r_{t+1} = \tilde r(\mathbf{s}_{t},a_t,\mathbf{s}_{t+1}).
\end{equation}

For the training sets of both CP benchmarks, the samples originate from trajectories of 100 (CPB) and 500 (CPSU) state transitions generated by a random walk on the benchmark dynamics.
The start states for these trajectories were sampled uniformly from $[-0.7,0.7]\times\{0\}\times[-2.4,2.4]\times \{0\}$ for CPB and from $[-\pi,\pi]\times\{0\}\times\{0\}\times\{0\}$ for CPSU.

We conducted several experiments to investigate the effect of different data set sizes and different network complexities. 
The results give a detailed impression about the minimum amount of data required to successfully apply the proposed FPSRL approach on different benchmarks and the adequate NN complexity for each data batch size. 
The experiments were conducted with network complexities of one, two, and three hidden layers with 10 hidden neurons each and arctangent activation functions.
For training, we used the Vario-Eta  algorithm~\citep{montavon:12}.
Training the networks can be executed in parallel and only requires a couple of minutes.
A detailed overview of the approximation performance of the resulting models, the FPSRL rules created with these models, and a comparison of non-interpretable policies generated by NFQ with the same data sets is given in Tables \ref{table:results_mc}, \ref{table:results_cpb}, and \ref{table:results_cpsu}. 
The mean squared errors of the normalized output variables (\textit{mean}=0, \textit{standard deviation}=1) have been depicted with respect to the generalization data sets.

\subsection{Policy Representations}
\label{subsection:fuzzy}

With the proposed FPSRL approach, we search for the parameterization $\mathbf{x}$ of a fuzzy policy formed by a certain number of rules.
The performance of an FPSRL policy is related to the number of rules because more rules generally allow a more sophisticated reaction to system states.
On the other hand, a higher number of rules requires more parameters to be optimized, which makes the optimizer's search problem more difficult.
In addition, a complex set of rules tends to be difficult or even impossible to interpret.
Thus, we determined that two rules are sufficient for the MC and CPB benchmarks, while adequate performance for the CPSU benchmark is only achievable with a minimum of four rules.
The output of the FPSRL policies is continuous, although a semi-discrete output can be obtained by increasing parameter $\alpha$ in Eq.~\eqref{eq:defuzzifier}.

We compared FPSRL policy training and its performance by applying NFQ to the same problems using the same data sets and approximative models.
NFQ was chosen because it is a well-established, widely applied, and well-documented RL methodology.
We used the NFQ implementation from the RL \textit{teachingbox}\footnote{Freely available at https://sourceforge.net/projects/teachingbox.} tool box.
In this paper, we did not aim to claim the proposed FPSRL approach is superior to the best RL algorithms in terms of performance; thus, NFQ was selected to show both the degree of difficulty of our benchmarks and the advantages and limitations of the proposed method.
Recent developments in deep RL have produced remarkable results with image-based online RL benchmarks~\citep{silver:14, vanhasselt:16}, and future studies may reveal that their performance with batch-based offline problems is superior to that of NFQ.
Nevertheless, these methods do not attempt to produce interpretable policies.
\section{Results}
\label{section:results}

\subsection{Mountain Car}

We performed 10 NFQ training procedures for the MC benchmark using the setup described in \ref{appendix:setup}. 
After each NFQ iteration, the latest policy was tested on the world model $\tilde{g}$ to compute an approximation $\mathcal{\tilde F}$ of the real performance. 
The policy yielding the best fitness value thus far was saved as an intermediate solution. 
To evaluate the true performance of the NFQ policies, we computed the true fitness value $\mathcal{F}$ by applying the policies to the mathematical MC dynamics $g$.

The difficulties in the MC benchmark are discontinuity in the velocity dimension when reaching the goal and the rather long horizon required to observe the effects of the applied actions.
With the first problem, it is difficult to model the goal area under the condition of limited samples reaching the goal using a random policy.
Training errors lead to a situation where the models do not correctly represent the state transitions at $\rho\geq0.6$, where the velocity suddenly becomes $\dot{\rho}=0$.
Rather, the models learn that $\rho\approx0.6$ yields $\dot{\rho}\approx0$.
Subsequently, during FPSRL training, the evaluation of policy candidates results in a situation where the car is driven to $\rho\approx0.6$ and is kept in this area by applying the correct forces, which leads to high reward transitions.
This problem could be solved by incorporating external knowledge about the goal area, which would result in a more convenient NN training process.
Here, we explicitly did not want to incorporate expert knowledge about the benchmarks. 
Instead, we wanted to demonstrate a purely data-driven autonomous learning example.
The results given in Table~\ref{table:results_mc} show that, despite these difficulties (even with small data batch sizes), well-performing policies can be learned using both FPSRL and NFQ.

For the MC benchmark and a discount factor $\gamma=0.9851$ (resulting from $q=0.05$), we consider a policy with performance $\mathcal{F}\approx-43$ or greater a successful solution to the benchmark problem. 
A policy with such performance can drive the car up the hill from any initial state in less than 200 time steps.

\begin{figure*}[tp]
    \centering
    \includegraphics[width=3.5in]{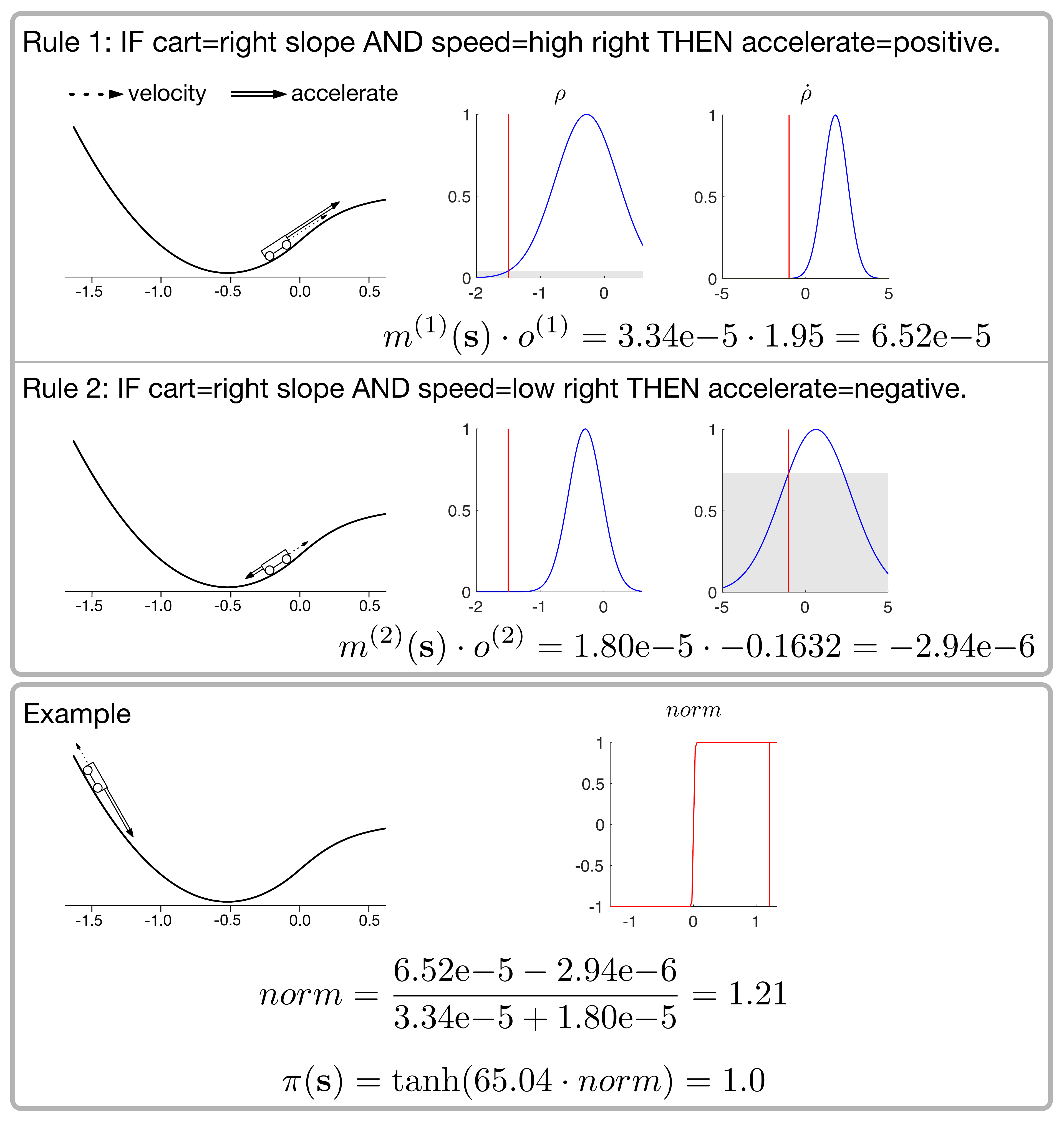}
    \caption{
    	Fuzzy rules for the MC benchmark (membership functions plotted in blue, example state position plotted in red with a gray area for the respective activation grade).
    	Both rules' activations are maximal at nearly the same position $\rho$, which implies that the $\rho$-dimension has minor influence on the policy's output.
    	This observation fits the fact that, for the MC benchmark, a simplistic but high-performing policy exists, i.e., accelerate the car in the direction of its current velocity.
    	Although this trivial policy yields good performance for the MC problem, better solutions exist.
    	For example, if you stop driving to the left earlier at a certain position, you reach the goal in fewer time steps, which yields a higher average return.
    	The depicted policy implements this advantageous solution as shown in the example section for state $\mathbf{s}=(-1.5,-1.0)$ .
    }
    \label{rules_mc}
\end{figure*}

One way to visualize fuzzy policies is to plot the respective membership functions and analyze the produced output for the sample states. 
A graphical representation of a policy for the MC benchmark is given in Fig.~\ref{rules_mc}. 
With some time for consideration, we were able to understand the policy's outputs for each considered state.

\begin{table*}[tp]
	\begin{center}
		\begin{tabular}{cccrrrrlrr} 
			\toprule
			Data & & \multicolumn{4}{c}{Models} & & \multicolumn{3}{c}{Policies}\\
			\cmidrule{1-1}\cmidrule{3-6}\cmidrule{8-10}				
			Batch size & & & \multicolumn{1}{c}{1 layer} & \multicolumn{1}{c}{2 layers} & \multicolumn{1}{c}{3 layers} & & & \multicolumn{1}{c}{FPSRL} & \multicolumn{1}{c}{NFQ} \\
			\cmidrule{1-1}\cmidrule{3-6}\cmidrule{8-10}
			1,000 & & $\rho$ & 4.67e-5 & 3.55e-5 & \textbf{3.05e-6}  &  & selected & \textbf{-41.98} &-43.23 \\
			& & $\dot{\rho}$ & 6.97e-3 & \textbf{3.54e-3} & 7.26e-3  &  & mean & -41.99 & -44.87 \\
			& & $r$ & 4.54e-1 & \textbf{1.46e-1} & 1.61e-1  &  & std & 0.01 & 1.33 \\
			\cmidrule{1-1}\cmidrule{3-6}\cmidrule{8-10}
			10,000 & & $\rho$ & 1.18e-5 & \textbf{3.34e-7} & 2.01e-6  &  & selected & \textbf{-42.22} &  -43.47\\
			& & $\dot{\rho}$ & 4.62e-3 & 3.48e-3 & \textbf{7.40e-5}  &  & mean & -42.69  & -45.73\\
			& & $r$ & 1.72e-2 & 2.54e-4 & \textbf{6.04e-7}  & & std & 0.46 & 2.90\\
			\cmidrule{1-1}\cmidrule{3-6}\cmidrule{8-10}
			100,000 & & $\rho$ & 1.55e-5 & \textbf{1.55e-7} & 2.88e-7  & & selected & \textbf{-41.99} & -43.12\\
			& & $\dot{\rho}$ & 1.10e-2  & 3.50e-4 & \textbf{5.15e-5}  & & mean & -41.93 & -43.28\\
			& & $r$ & 1.01e-3 & 2.09e-6 & \textbf{5.85e-8}  & & std & 0.11 & 1.22\\
			\bottomrule	
		\end{tabular}
		\caption{
			MC results (left to right): 
			(1) data: number of state transitions, obtained from random trajectories on the benchmark dynamics; 
			(2) models: generalization errors of the best NN models we were able to produce given a certain amount of data and pre-defined network complexity; 
			(3) policies: performance with the real benchmark dynamics of different policy types trained/selected according to the performance using the models from the left.
			For each policy setting, 10 training experiments were performed to obtain statistically meaningful results.
			The presented results for different data batch sizes show that the MC benchmark dynamics are rather easy to model using NNs.
			In addition, models having significantly greater errors with the generalization sets were still sufficient for training a fuzzy policy using FPSRL and selecting a well-performing policy from NFQ.
		}
		\label{table:results_mc}
	\end{center}
\end{table*}

\subsection{Cart-pole Balancing}

The CPB benchmark has two different discontinuities in its dynamics, which make the modeling process more difficult compared to the MC benchmark case.
The first discontinuity occurs when the cart leaves the restricted state space and ends up in the failure state, i.e., as soon as $\abs{\theta}>0.7$ or $\abs{\rho}>2.4$, the cart becomes fixed at its current position (both velocities $\dot{\theta}$ and $\dot{\rho}$ become zero).
The second discontinuity appears when the cart enters the goal region.
In this region, the reward switches from $-0.1$  to $0.0$, which is a rather small change compared to the failure state reward of $-1.0$.
In addition to the difficulty in modeling discrete changes with NNs, this task becomes even more complicated if samples for these transitions are rare.
In contrast to the difficulties in modeling the benchmark dynamics, a rather simple policy can balance the pole without leaving the restricted state space.
With the discount factor $\gamma=0.97$ ($q=0.05$), we consider policies that yield a performance of $\mathcal{F}\approx-1.5$ or greater as successful. 

The task for FPSRL was to find a parameterization for two fuzzy rules. 
Here, we used 100 particles and an out-of-the-box PSO setup (\ref{appendix:setup}). 
The training employed 1,000 start states that were uniformly sampled from $[-0.5,0.5]\times\{0\}\times [-0.5,0.5]\times\{0\}$ (Table~\ref{table:results_cpb}).
Note that a data batch size of 100,000 sample transitions was required to build models with adequate approximation quality for training a model-based RL policy.
Models trained with 1,000 or 10,000 sample transitions could not correctly approximate the effects that occurred when entering the failure-state area.
Further, they incorrectly predicted possibilities to escape the failure state and to balance the pole in subsequent time steps.
The model-based FPSRL technique exploited these weaknesses and produced policies that perform well with the models but demonstrated poor performance with the real benchmark dynamics.

\begin{figure*}[tp]
    \centering
    \includegraphics[width=5.0in]{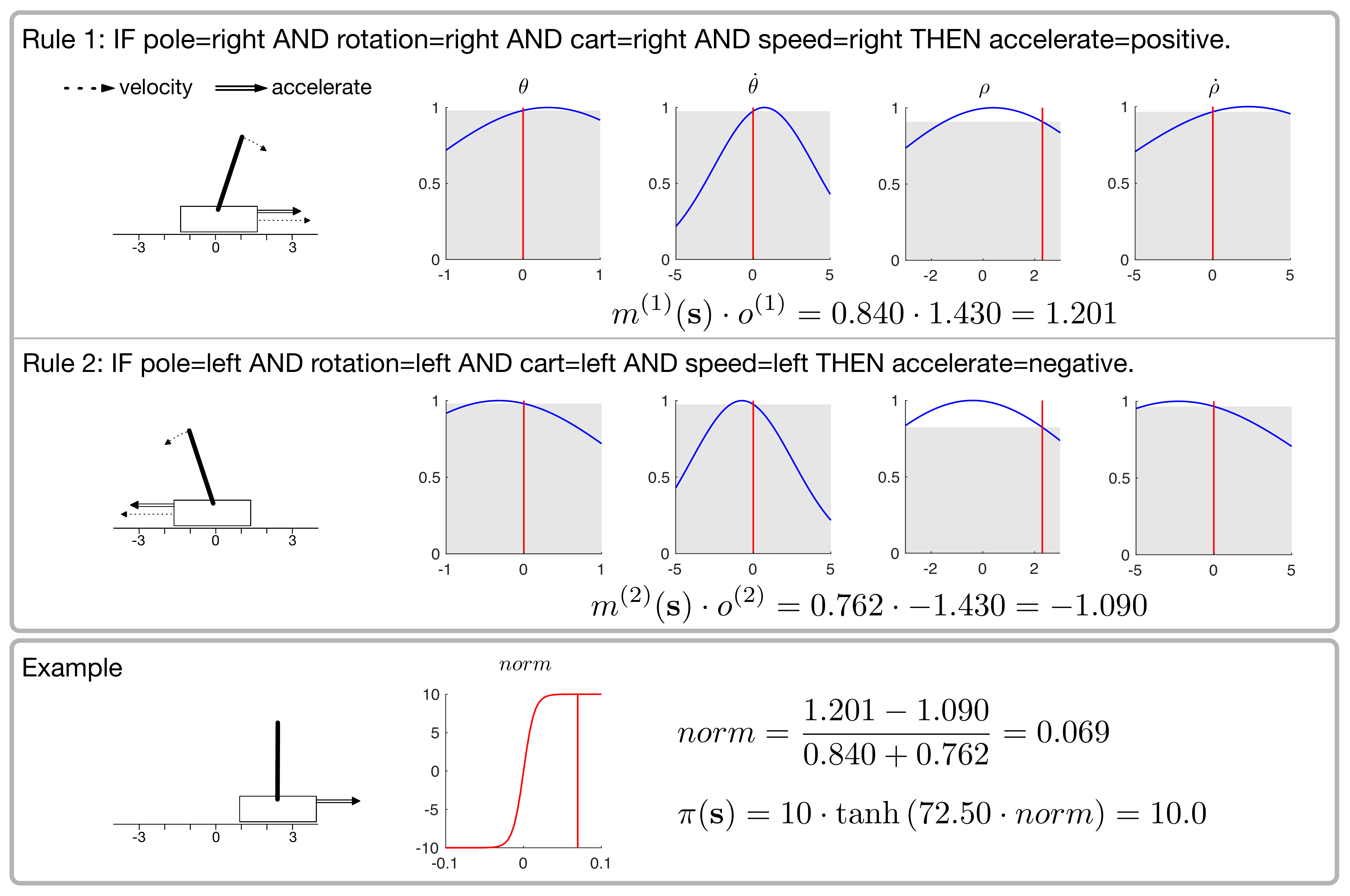}
    \caption{
    	Fuzzy rules for the CPB benchmark.
    	The visualization of a fuzzy policy can be useful for revealing the weaknesses of a given set of rules.
    	For example, despite the presented policy being the result of successful FPSRL training and yielding high average returns for all test states, states at the boundary of the allowed cart position would result in failed episodes.
    	The depicted example shows that state $\mathbf{s}=(0,0,2.3,0)$ would activate rule 1; thus, it would accelerate the car even further positive and eventually end up in a failure state.
    	Therefore, by examining only the rule set, the elementary weaknesses of the policy can be identified and the test state set can be adapted appropriately.
    }
    \label{rules_cpreg}
\end{figure*}

A visual representation of one of the resulting fuzzy policies is given in Fig.~\ref{rules_cpreg}.
This example illustrates a situation where the potential problems of a policy can be observed via visual inspection, which is a significant advantage of interpretable policies.

In contrast to FPSRL, NFQ could produce well-performing non-interpretable policies even with small data batch sizes.
Note that the same weak models used for FPSRL training were used to determine which NFQ iteration produced the best policy during NFQ training with 1,000 episodes.
In our experiments, we observed that even models with bad approximative quality when simulating the benchmark dynamics are useful for NFQ policy selection because NFQ training never observed the models during training and therefore was not prone to exploiting their weaknesses.

\begin{table*}[tp]
	\begin{center}
		\begin{tabular}{cccrrrrlrr} 
			\toprule
			Data & & \multicolumn{4}{c}{Models} & & \multicolumn{3}{c}{Policies}\\
			\cmidrule{1-1}\cmidrule{3-6}\cmidrule{8-10}	
			Batch size & & & \multicolumn{1}{c}{1 layer} & \multicolumn{1}{c}{2 layers} & \multicolumn{1}{c}{3 layers} & & & \multicolumn{1}{c}{FPSRL} & \multicolumn{1}{c}{NFQ} \\
			\cmidrule{1-1}\cmidrule{3-6}\cmidrule{8-10}
			1,000 & & $\theta$ & 1.57e-7 & 1.37e-7 & \textbf{1.07e-7}  &  & selected & -9.03  & \textbf{-1.35} \\
			& & $\dot{\theta}$ & 4.62e-2 & 6.03e-2 & \textbf{8.51e-3}  &  & mean & -14.59 & -1.82 \\
			& & $\rho$ & \textbf{4.32e-8} & 8.29e-8 & 1.29e-7  &  & std & 5.34 & 0.54 \\
			& & $\dot{\rho}$ & 4.33e-2 & \textbf{1.33e-2} & 1.03e-1  &  & &  &\\
			& & $r$ & 2.09e-2 & 2.58e-2 & \textbf{1.11e-2}  &  & &  &\\
			\cmidrule{1-1}\cmidrule{3-6}\cmidrule{8-10}
			10,000 & & $\theta$ & \textbf{5.95e-9} & 3.79e-8 & 2.84e-8  & & selected & -3.29  & \textbf{-0.99} \\
			& & $\dot{\theta}$ & 3.68e-2 & 1.07e-2 & \textbf{5.08e-3}  &  & mean & -3.30 & -1.18 \\
			& & $\rho$ & \textbf{9.98e-9} & 8.12e-7 & 4.82e-8  &  & std & 0.02 & 0.23\\
			& & $\dot{\rho}$ & 5.18e-2 & 4.16e-2 & \textbf{4.02e-2}  &  & &  &\\
			& & $r$ & 1.22e-2 & \textbf{4.75e-4} & 6.17e-4  &  & & &\\
			\cmidrule{1-1}\cmidrule{3-6}\cmidrule{8-10}
			100,000 & & $\theta$ & \textbf{5.73e-9} & 2.43e-8 & 2.69e-8  &  & selected & \textbf{-1.31} & -1.81\\
			& & $\dot{\theta}$ & 3.55e-2 & 1.25e-2 & \textbf{9.93e-3}  &  & mean & -1.31 & -2.03\\
			& & $\rho$ & \textbf{2.91e-8} & 3.44e-8 & 1.41e-7  &  & std & 8.97e-4 & 0.24\\
			& & $\dot{\rho}$ & 2.83e-2 & 2.43e-2 & \textbf{1.30e-2}  &  & &  &\\
			& & $r$ & 5.86e-3 & \textbf{1.08e-4} & 9.03e-4  &  & &  &\\
			\bottomrule	
		\end{tabular}
		\caption{
			Cart-pole balancing results. 
			The experiments show that the modeling process of variables containing nonlinearities is difficult and requires an adequate amount of sample data. 
			Since both the pendulum and the cart velocities suddenly become zero if the failure state is reached, the modeling process requires a certain number of these events in the training data to correctly model this effect.
			The results for a batch size of 1,000 show that a model that is not applicable to model-based RL can still be used for policy selection for a model-free RL technique such as NFQ.
			As the models' errors reduce with increasing data batch sizes, FPSRL becomes increasingly capable of finding well-performing interpretable policies.
			Note that we encountered an effect that reduced NFQ performance can occur even if the data batch size increases.
		}
		\label{table:results_cpb}
	\end{center}
\end{table*}

\subsection{Cart-pole Swing-up}

Compared to the MC and CPB benchmarks, the results for the CPSU benchmark show a completely different picture in terms of performance and the training process.
Despite CPB and CPSU sharing the same underlying mathematical transition dynamics, they differ in the following two important aspects.
First, discontinuities in the state transitions do not occur owing to the absence of a failure state area.
Second, the planning horizon for a successful policy is significantly higher.
While the latter makes it particularly difficult to find a solution by applying standard NFQ, the first property makes CPSU a good example of the strength of the proposed FPSRL approach.

NFQ's performance decreased dramatically for the CPSU problem (Table~\ref{table:results_cpsu}).
For this benchmark with $\gamma=0.994$ ($q=0.05$), solutions with $\mathcal{F}\approx50$ or greater with a set of 1,000 benchmark states uniformly sampled from $[-\pi,\pi]\times\{0\}\times [-0.5,0.5]\times\{0\}$ were considered successful. 
The policies exhibiting such performance can swing-up more than $99\%$ of the given test states.
In our experiments, none of the NFQ trainings produced such a successful policy.

In contrast, the proposed FPSRL could find a parameterization for successful policies using four fuzzy rules by assessing their performance on world models trained with data batch sizes of 10,000 or greater.
For a data batch size of 1,000, the transition samples containing the goal area reward were far too few in the data set to model this area correctly.
However, the extremely high errors obtained with the generalization set during model training are excellent indicators of this weakness.

\begin{figure*}[tp]
	\centering
	\includegraphics[width=5.0in]{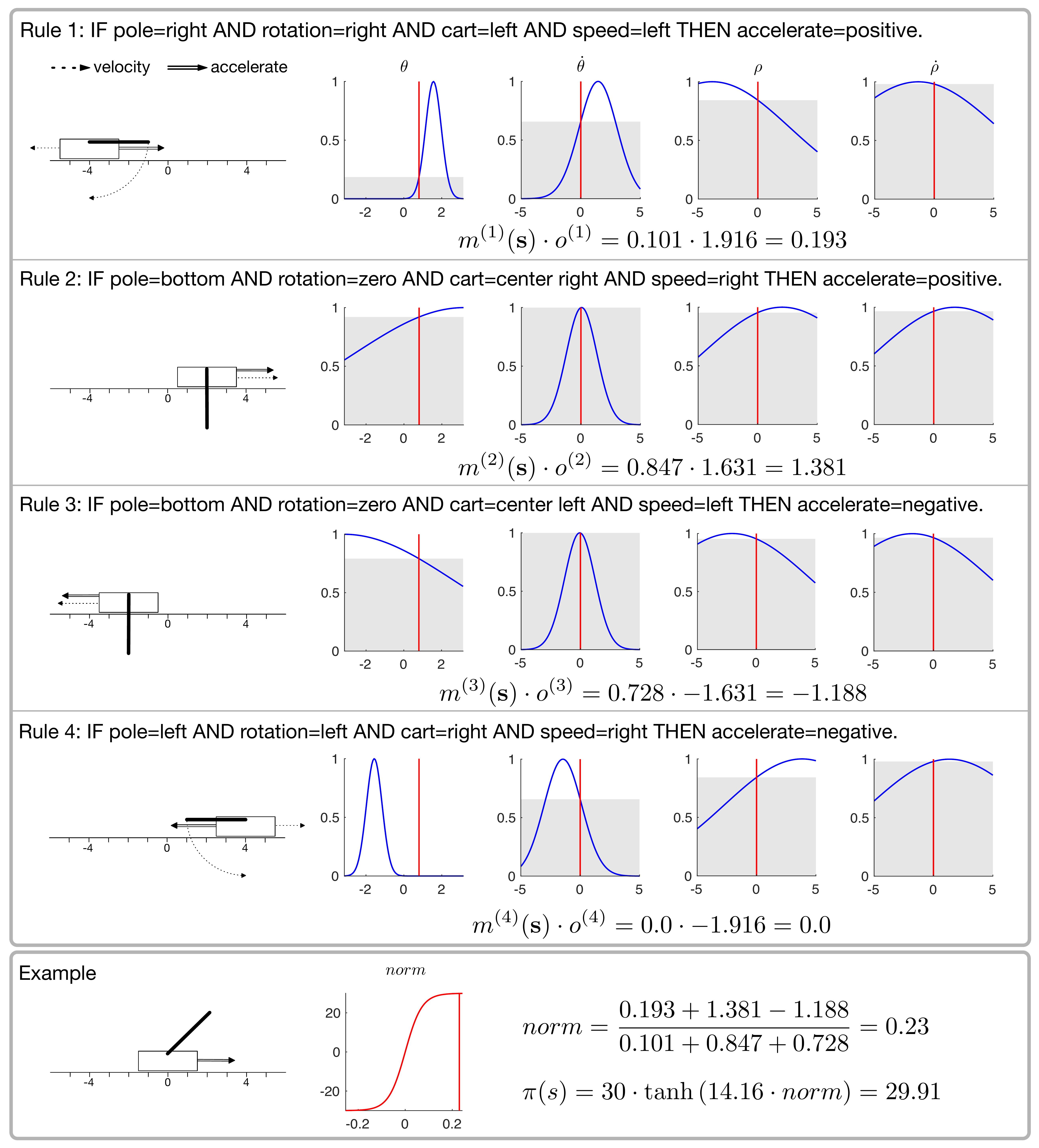}
	\caption{
		Fuzzy rules for CPSU benchmark.
		Even with four rules, fuzzy policies can be visualized in an easily interpretable way.
		By inspecting the prototype cart-pole diagrams for each rule, two basic concepts can be identified for accelerating in each direction.
		First (Rule 1 (4)): the cart's position is on the left (right) and moving further to the left (right), while the pole is simultaneously falling on the right (left). Then the cart is accelerated towards the right (left).
		Second (Rule 2 (3)): the cart is between the center and the right (left) and the pole is hanging down. Then the cart is accelerated towards the right (left).
		Both prototypes are utilized to realize the complex task of swinging the pole up.
		Balancing of the pole while the cart is centered around $\rho=0$ is realized via fuzzy interaction of these prototype rules, as shown in the example in the last row.
	}
	\label{rules_cpsu}
\end{figure*}

Figure~\ref{rules_cpsu} shows how even more complex fuzzy policies can be visualized and help make RL policies interpretable.

\begin{table*}[tp]
	\begin{center}
		\begin{tabular}{cccrrrrlrr} 
			\toprule
			Data & & \multicolumn{4}{c}{Models} & & \multicolumn{3}{c}{Policies}\\
			\cmidrule{1-1}\cmidrule{3-6}\cmidrule{8-10}	
			Batch size & & & \multicolumn{1}{c}{1 layer} & \multicolumn{1}{c}{2 layers} & \multicolumn{1}{c}{3 layers} & & & \multicolumn{1}{c}{FPSRL} & \multicolumn{1}{c}{NFQ} \\
			\cmidrule{1-1}\cmidrule{3-6}\cmidrule{8-10}	
			1,000 & & $\theta$ & 2.02e-4 & \textbf{2.61e-6} & 3.07e-6  &  & selected  & -157.49 & \textbf{-153.59}\\
			& & $\dot{\theta}$ & 2.93e-3 & \textbf{4.65e-4} & 5.78e-4  &  & mean  & -156.53 & -156.43\\
			& & $\rho$ & 2.27e-5 & \textbf{1.44e-5} & 1.85e-5  &  & std  & 2.30 &  1.64\\
			& & $\dot{\rho}$ & 9.85e-4 & \textbf{9.90e-5} & 1.13e-3  &   & &  & \\
			& & $r$ & \textbf{5.00} & 5.07 & 5.06  & & & & \\
			\cmidrule{1-1}\cmidrule{3-6}\cmidrule{8-10}	
			10,000 & & $\theta$ & 3.23e-6 & \textbf{2.17e-6} & 2.31e-6  &  & selected  & \textbf{-34.03}  & -134.82 \\
			& & $\dot{\theta}$ & 9.86e-5 & \textbf{7.88e-}5 & 3.65e-4  &  & mean & -53.82 & -153.63 \\
			& & $\rho$ & 3.06e-6 & \textbf{1.48e-6} & 2.08e-6  &  & std  & 12.01 & 6.69\\
			& & $\dot{\rho}$ &1.13e-5 & \textbf{8.83e-6} & 3.39e-5  &   & &   & \\
			& & $r$ & 1.68e-1 & \textbf{5.05e-2} & 9.42e-2  &   & &   & \\
			\cmidrule{1-1}\cmidrule{3-6}\cmidrule{8-10}	
			100,000 & & $\theta$ & \textbf{2.00e-6} & 3.07e-6 & 2.56e-6  &   & selected  & \textbf{-32.42} & -150.93\\
			& & $\dot{\theta}$ & \textbf{2.63e-4} & 5.62e-4 & 3.38e-4  & & mean & -53.22 &  -152.66\\
			& & $\rho$ & \textbf{8.83e-6} & 1.23e-5 & 4.81e-5  &   &std &  11.17 & 2.26\\
			& & $\dot{\rho}$ & 2.47e-5 & 6.46e-5 & \textbf{2.28e-5}  &   & &   & \\
			& & $r$ & 1.95e-1 & \textbf{4.76e-3} & 7.77e-3  &   & &   & \\
			\bottomrule	
		\end{tabular}
		\caption{
			Cart-pole swing-up results.
			High errors with the generalization set when training the reward function with a data batch size of less than 10,000 clearly indicate the absence of an adequate number of transition samples that describe the effects noted when reaching the goal area.
			Smooth and easy dynamics in the other dimensions make it rather easy to model the CPSU dynamics and subsequently use them to conduct model-based RL.
			Note that the long planning time horizon required in this benchmark made it impossible to learn successful policies with the standard NFQ.
		}
		\label{table:results_cpsu}
	\end{center}
\end{table*}
\section{Conclusion}
\label{chapter:Conclusion}

The traditional way to create self-organizing fuzzy controllers either requires an expert-designed fitness function according to which the optimizer finds the optimal controller parameters or relies on detailed knowledge regarding the optimal controller policy. 
Either requirement is difficult to satisfy when dealing with real-world industrial problems. 
However, data gathered from the system to be controlled using some default policy are available in many cases. 

The FPSRL approach proposed herein can use such data to produce high-performing and interpretable fuzzy policies for RL problems.
Particularly for problems where system dynamics are rather easy to model from an adequate amount of data and where the resulting RL policy can be expected to be compact and interpretable, the proposed FPSRL approach might be of interest to industry domain experts.

The experimental results obtained with three standard RL benchmarks have demonstrated the advantages and limitations of the proposed model-based method compared with the well-known model-free NFQ approach.
However, the results obtained with the CPB problem reveal an important limitation of FPSRL, i.e.,  training using weak approximation models.
The proposed approach can exploit these weaknesses, which can potentially result in poor performance when evaluated using the real dynamics.
Modeling techniques that can provide a measure of uncertainty in their predictions, such as Gaussian processes or Bayesian NNs, can possibly overcome these problems.
Recent developments in modeling stochastic dynamic systems~\citep{depeweg:16} may provide an approximation of the mean of the next system state but also compute uncertainty for transitions in the state-action space.

In addition, continuous state and action spaces, as well as long time horizons, do not appear to introduce obstacles to the training of fuzzy policies.
The resulting policies obtained with the CPSU benchmark performed significantly better than those generated by the standard NFQ.

However, one of the most significant advantages of the proposed method over other RL methods is the fact that fuzzy rules can be easily and conveniently visualized and interpreted.
We have suggested a compact and informative approach to present fuzzy rule policies that can serve as a basis for discussion with domain experts.

The application of the proposed FPSRL approach in industry settings could prove to be of significant interest because, in many cases, data from systems are readily available and interpretable fuzzy policies are favored over black-box RL solutions, such as Q-function-based model-free approaches.
\section*{Acknowledgment}

The project this report is based on was supported with funds from the German Federal Ministry of Education and Research under project number 01IB15001. 
The sole responsibility for the report's contents lies with the authors.

The authors would like to thank Dragan Obradovic and Clemens Otte for their insightful discussions and helpful suggestions.

\section*{References}

\bibliographystyle{elsarticle-harv}
\biboptions{longnamesfirst,authoryear,round,semicolon}
\bibliography{paper}

\appendix
\section{PSO Algorithm}
\label{appendix:algorithm}

Algorithm~\ref{pso_alg} explains in pseudocode the PSO algorithm applied in our experiments.

\begin{algorithm}
\begin{small}
\begin{singlespace}
    \KwData{
        \begin{itemize}
            \setlength{\parskip}{0pt}
            \item $N$ randomly initialized $d$-dimensional particle \\
            positions with \\
            	$\mathbf{x}_i=\mathbf{y}_i$ and velocities $\mathbf{v}_i$ of particle $i$, where $i=1,\ldots,N$
            \item Fitness function $\mathcal{F}$ (Eq.~\eqref{eq:fitness_function})
            \item Inertia weight factor $w$ and acceleration \\
            constants $c_1$ and $c_2$
            \item Random number generator rand()
            \item Search space boundaries $\mathbf{x}_{\text{min}}$ and $\mathbf{x}_{\text{max}}$
            \item Velocity boundaries $\mathbf{v}_{\text{min}}=-0.1\cdot(\mathbf{x}_{\text{max}}-\mathbf{x}_{\text{min}})$ \\
            and $\mathbf{v}_{\text{max}}=0.1\cdot(\mathbf{x}_{\text{max}}-\mathbf{x}_{\text{min}})$
            \item Swarm topology graph defining neighborhood $\mathcal{N}_i$
        \end{itemize}
    }
    \KwResult{
        \begin{itemize}
            \item Global best position $\hat{\mathbf{y}}$
        \end{itemize}
    }
    \SetAlgoLined
    \Repeat{Stopping criterion is met}{
        \ForEach{Particle $i$}{
            $\sslash$ Neighborhood best position of\\
            $\sslash$ particle $i$ (Eq.~\eqref{neighborhood_best})\;
            $\hat{\mathbf{y}}_i\leftarrow\argmax_{\mathbf{z}\in\{\mathbf{y}_j\, \mid\, j\in\mathcal N_i\}}\mathcal{F}(\mathbf{z})$\;

        }
        $\sslash$ Position updates\;
        \ForEach{Particle $i$}{
            $\sslash$ Determine new velocity of particle $i$ (Eq.~\eqref{basic_pso})\;
            \For{$j=1,\ldots,d$}{
                $v_{ij}\leftarrow wv_{ij}+c_1\cdot\text{rand()}\cdot[y_{ij}-x_{ij}]+c_2\cdot\text{rand()}\cdot[\hat{y}_{ij}-x_{ij}]$\;
            }
            $\sslash$ Truncate particle $i$'s velocity\;
            \For{$j=1,\ldots,d$}{
                $v_{ij}\leftarrow\min(v_{{\text{max}}_j},\max(v_{{\text{min}}_j},v_{ij}))$
            }
            $\sslash$ Compute new position of particle $i$ (Eq.~\eqref{position_update})\;
            $\mathbf{x}_i\leftarrow \mathbf{x}_i+\mathbf{v}_i$\;
            $\sslash$ Truncate particle $i$'s position\;
            \For{$j=1,\ldots,d$}{
                $x_{ij}\leftarrow\min(x_{{\text{max}}_j},\max(x_{{\text{min}}_j},x_{ij}))$
            }            
            $\sslash$ Personal best positions (Eq.~\eqref{personal_best})\;
            \If{$\mathcal{F}(\mathbf{x}_i)>\mathcal{F}(\mathbf{y}_i)$}{
                $\sslash$ Set new personal best position of particle $i$\;
                $\mathbf{y}_i\leftarrow \mathbf{x}_i$\;
            }
        }
    }
    $\sslash$ Determine the global best position\;
    $\hat{\mathbf{y}}\leftarrow\argmax_{\mathbf{z}\in\{\mathbf{y}_1,\ldots,\mathbf{y}_N\}}\mathcal{F}(\mathbf{z})$\;
    \Return{$\hat{\mathbf{y}}$}
    \caption{PSO algorithm. Particle $i$ is represented by position $\mathbf{x}_i$, personal best position $\mathbf{y}_i$, and neighborhood best position $\hat{\mathbf{y}}_i$.}
    \label{pso_alg}
    \end{singlespace}
    \end{small}
\end{algorithm}

\section{Experimental Setup}
\label{appendix:setup}

Table~\ref{table:setup} gives a compact overview of the parameters used for the experiments presented herein.
Note that extensive parameter studies for both FPSRL and NFQ are beyond the scope of this paper.
Nevertheless, we evaluated various parameters known from the literature or our own experience, and we noted that the presented setup was the most successful of all the setups tested in our experiments.

\begin{table*}[tp]
	\begin{center}
		\begin{tabular}{lccc} 
			\toprule
			& MC & CPB & CPSU \\
			\midrule			
			Benchmark & & & \\
			\qquad State dimensionality $D$ & 2 & 4 & 4\\
			\qquad Time horizon $T$ & 200 & 100 & 500\\
			\qquad Discount factor $\gamma$ & 0.9851 & 0.9700 & 0.9940\\
			\midrule	
			FPSRL & & & \\
			\qquad Number of particles $N$ & 100 & 100 & 1,000\\
			\qquad PSO iterations & 1,000 & 1,000 & 1,000\\
			\qquad PSO topology & ring & ring & ring\\
			\qquad Number of rules $C$ & 2 & 2 & 4\\
			\qquad Rule parameters $\lvert\mathbf{x}\rvert$ & 11 & 10 & 19\\
			\qquad Actions & $[-1,1]$ & $[-10,10]$ & $[-30,30]$\\
			\midrule	
			NFQ & & & \\
			\qquad Q iterations & 1,000 & 1,000 & 1,000\\
			\qquad NN epochs & 300 & 300 & 300\\
			\qquad NN layers & 3-20-20-1 & 5-20-20-1 & 5-20-20-20-1\\
			\qquad NN activation & sigmoid & sigmoid & sigmoid\\
			\qquad Actions & $\{-1,0,1\}$ & $\{-10,0,10\}$ & $\{-30,0,30\}$\\
			\bottomrule	
		\end{tabular}
		\caption{
			Experimental setup.
		}
		\label{table:setup}
	\end{center}
\end{table*}

\end{document}